\documentclass[runningheads]{llncs}
\usepackage[T1]{fontenc}
\usepackage{graphicx}
\usepackage{amsmath}
\usepackage{amssymb}
\usepackage{multirow}
\usepackage{array}
\newcolumntype{C}[1]{>{\centering\arraybackslash}p{#1}}

\begin{document}
\title{PU-UNet: Stable Multiplicative Interactions for Medical Image Segmentation}
%

\author{Ziyuan Li\inst{1,2} \and Osamah Sufyan\inst{1} \and
Uwe Jaekel\inst{1} \and
Babette Dellen\inst{1}}

\institute{Department of Mathematics, Informatics and Technology, University of Applied Sciences Koblenz, Joseph-Rovan-Allee 2, 53424 Remagen, Germany \and
Technical University of Munich, Munich, Germany\\ \email{ziyuan.li@tum.de}}

\maketitle              
\begin{abstract}
Many dense prediction networks rely on additive feature transformations and model higher-order feature interactions only implicitly. Product units provide an explicit mechanism for multiplicative feature modeling, but their logarithmic--exponential formulation can cause numerical instability, which has limited their use in deep dense prediction networks. In this work, we propose Product-Unit U-Net (PU-UNet), a residual U-Net that integrates stable product-unit residual blocks into rich low-resolution stages for medical image segmentation. The proposed formulation combines smooth positivity mapping with log-domain clipping, enabling stable multiplicative feature learning with negligible computational overhead. On ISIC 2018, Kvasir-SEG, and BUSI, PU-UNet achieves Dice scores of 0.942, 0.959, and up to 0.925, respectively. Compared with a matched Residual U-Net baseline, PU-UNet consistently improves Dice and IoU while keeping parameters, FLOPs, and inference latency nearly unchanged, and reduces the image-level false-positive rate on normal BUSI cases from 0.077 to zero. Ablation studies suggest that the gains are associated with product-unit interactions, are strongest under low-resolution placement, and benefit from the proposed stabilization design. These results suggest that stable product-unit residual learning can be an effective way to enhance U-Net-style segmentation networks with explicit multiplicative interactions.
\end{abstract}

\section{Introduction}
Medical image segmentation is a core task in medical image analysis, where U-Net and its variants have become dominant due to their encoder--decoder structure, skip connections, and strong performance across biomedical imaging tasks~\cite{ronneberger2015u,milletari2016v,oktay2018attention,isensee2021nnu}. However, most U-Net-style models rely mainly on additive feature transformations, in which convolutional responses are aggregated through weighted sums followed by pointwise nonlinearities. Such architectures model complex cross-feature dependencies only implicitly, which may be suboptimal when segmentation depends on nonlinear interactions among texture, shape, boundary, and contrast cues~\cite{dellen2019function,li2025deep}.

Product units offer a direct way to model such interactions by introducing multiplicative feature combinations~\cite{durbin1989product,leerink1995learning,dellen2019function}. They provide an inductive bias for higher-order nonlinear dependencies and have shown promise in several settings~\cite{dellen2019function,li2025deep,li2025advancing}. Nevertheless, integrating product units into deep dense prediction networks remains challenging because the logarithmic--exponential formulation is prone to numerical instability and optimization issues~\cite{engelbrecht1999training,dellen2019function}. As a result, despite their long-standing theoretical appeal, product units have only rarely been used in deeper neural networks, and their potential for modern medical image segmentation remains underexplored.

In this work, we show that product units can serve as an effective and efficient component for medical image segmentation when integrated into U-Net appropriately. We propose Product-Unit U-Net (PU-UNet), a Residual U-Net that introduces explicit multiplicative interactions through stable product-unit residual blocks. Rather than replacing standard convolutions throughout the network, we insert product-unit blocks selectively into semantically richer low-resolution stages, where multiplicative modeling is expected to be more useful than in shallow high-resolution layers dominated by local appearance cues~\cite{zeiler2014visualizing,long2015fully}. To make this design practical, we stabilize the product-unit operation using smooth positivity mapping and log-domain clipping before exponentiation.

We evaluate PU-UNet on ISIC 2018~\cite{codella2019skin}, Kvasir-SEG~\cite{jha2019kvasir}, and BUSI~\cite{al2020dataset}, covering skin lesion, polyp, and breast ultrasound segmentation. Across all datasets, PU-UNet consistently improves over a matched Residual U-Net baseline with almost unchanged parameter count, FLOPs, and inference latency. On BUSI, it also reduces false positives on normal cases, highlighting its practical value in a challenging ultrasound setting.

The main contributions of this work are threefold:
(i) we develop a stable product-unit formulation with smooth positivity mapping and log-domain clipping, addressing a key numerical barrier that has limited the use of product units in deeper networks;
(ii) we propose PU-UNet, a Residual U-Net that introduces product-unit residual blocks into semantically richer low-resolution stages;
(iii) we demonstrate on ISIC 2018, Kvasir-SEG, and BUSI that explicit multiplicative modeling consistently improves a matched Residual U-Net baseline with negligible overhead, and that selective low-resolution insertion is more effective than full-network replacement.

\section{Background on Product Units}
A product unit (PU) explicitly models multiplicative interactions among input features. For a positive input $x=(x_1,\dots,x_d)$, it is defined as
\begin{equation}
y=\prod_{i=1}^{d} x_i^{w_i}
=\exp\left(\sum_{i=1}^{d} w_i \log x_i\right),
\label{eq:pu_basic}
\end{equation}
where $w$ are learnable weights~\cite{dellen2019function}. Unlike standard summation-based neurons, PUs provide an explicit inductive bias for higher-order feature interactions.

For image data, this idea can be applied locally over convolutional receptive fields. Given a feature map $X$, a 2D product unit at spatial location $(i,j)$ is written as
\begin{equation}
\mathrm{ConvPU}(X)=
\exp\left(\mathrm{Conv}(\log X)\right)
=\exp\left(\sum_{m,n} W(m,n)\log X(i+m,j+n)\right),
\label{eq:pu2d}
\end{equation}
where $W$ is the learnable kernel~\cite{li2025deep}.

Product units can also be embedded into residual blocks to improve trainability~\cite{li2025deep}. In this formulation, a conventional convolution is followed by a 2D product-unit transformation:
\begin{equation}
Z=\mathrm{Conv}(X), \qquad
Y = X + \mathrm{ConvPU}(\tilde{Z}),
\label{eq:pure_block}
\end{equation}
where the PU input is stabilized by a learnable lower bound,
\begin{equation}
\tilde{Z}=\max\bigl(Z,\mathrm{softplus}(\theta)+\varepsilon\bigr),
\label{eq:pu_clip}
\end{equation}
with trainable parameter $\theta$ and a small constant $\varepsilon>0$. This guarantees positivity before the logarithm, but the subsequent exponential reconstruction can still lead to large activation ranges in dense prediction networks, motivating the stabilized formulation introduced below.

\section{Method}
Given an input image $X \in \mathbb{R}^{C \times H \times W}$, PU-UNet predicts a pixel-wise segmentation logit map $\hat{Y} \in \mathbb{R}^{1 \times H \times W}$. The network follows a Residual U-Net architecture with an encoder--decoder backbone and skip connections, while selected residual blocks are replaced by product-unit residual blocks. This design introduces explicit multiplicative feature interactions without sacrificing numerical stability.

\subsection{Stable 2D Product-Unit}
\begin{figure}[b]
    \centering
    \includegraphics[width=\linewidth]{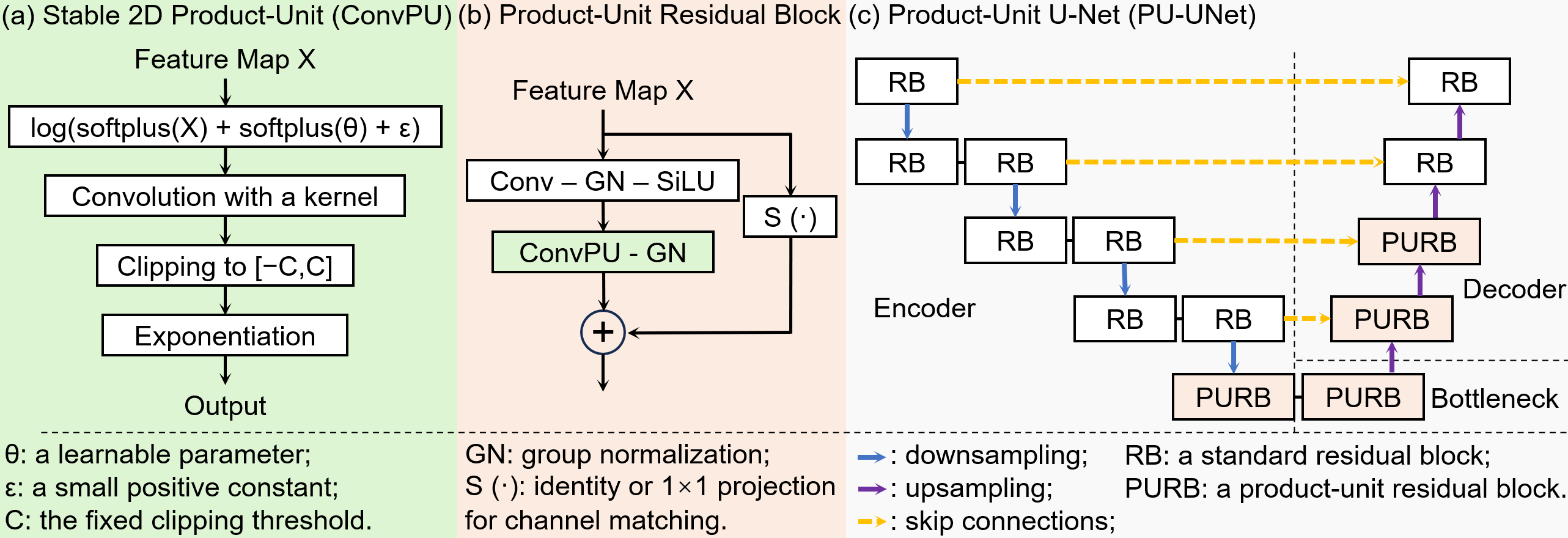}
    \caption{Overview of the proposed PU-UNet, including the network architecture and core product-unit components.}
    \label{fig:net}
\end{figure}

Earlier image-domain product-unit formulations mainly ensure that the logarithm is well-defined by enforcing a positive lower bound for the input~\cite{li2025deep}. However, in dense prediction, feature magnitudes can vary substantially across spatial locations and training iterations, so the exponential reconstruction in Eq.~\eqref{eq:pu2d} may still become numerically unstable. We therefore introduce a more robust 2D product-unit formulation for segmentation. The computation flow of the proposed operator is illustrated in Fig.~\ref{fig:net}(a).

Given an input feature map $X$, we first map it to a strictly positive domain by
\begin{equation}
X^{+} = \mathrm{softplus}(X) + \theta,
\qquad
\theta = \mathrm{softplus}(\theta_{\mathrm{raw}}) + \varepsilon,
\label{eq:method_pos}
\end{equation}
where $\theta_{\mathrm{raw}}$ is a learnable scalar shared across channels and spatial locations, and $\varepsilon>0$ is a small constant ($10^{-7}$). Compared with hard lower-bound truncation (Eq.~\eqref{eq:pu_clip})~\cite{li2025deep}, this yields a smooth positive-domain mapping and avoids a non-smooth clipping boundary before the logarithmic transformation. We then compute
\begin{equation}
L = \mathrm{Conv}\!\bigl(\log X^{+}\bigr),
\label{eq:method_logconv}
\end{equation}
followed by an explicit clamp in the log domain with threshold $C>0$:
\begin{equation}
\bar{L} = \mathrm{clip}(L,-C,C).
\label{eq:method_clip}
\end{equation}
In all experiments, we set the log-domain clipping threshold to $C=10$, which keeps the exponential reconstruction within the FP16 range under automatic mixed precision training, since $\exp(10)\approx 2.2\times 10^{4}$ is below the FP16 maximum value ($\approx 6.55\times 10^{4}$). The PU response is finally reconstructed as
\begin{equation}
\mathrm{ConvPU}(X)=\exp(\bar{L}).
\label{eq:method_convpu}
\end{equation}
By bounding the log-domain response before exponentiation, the proposed formulation controls the dynamic range of the exponential and reduces overflow and non-finite activations during training.

\subsection{Product-Unit Residual Block}
Our segmentation backbone uses a product-unit residual U-Net block with group normalization (GN) and SiLU activation~\cite{elfwing2018sigmoid}. Unlike earlier product-unit residual formulations (Eqs.~\eqref{eq:pure_block} and~\eqref{eq:pu_clip}), we keep the first \texttt{Conv--GN--SiLU} stage of the baseline residual block and replace only the second convolution with the stable 2D product-unit operator in Eq.~\eqref{eq:method_convpu}, as illustrated in Fig.~\ref{fig:net}(b). Let
\begin{equation}
H=\mathrm{SiLU}\!\bigl(\mathrm{GN}(\mathrm{Conv}(X))\bigr),
\end{equation}
then the proposed product-unit residual block is defined as
\begin{equation}
\mathcal{R}_{\mathrm{PU}}(X)=S(X)+\mathrm{GN}\!\bigl(\mathrm{ConvPU}(H)\bigr),
\label{eq:method_puresblock}
\end{equation}
where $S(\cdot)$ is the identity shortcut or a $1\times1$ projection for channel matching.

This design preserves the basic processing pattern of the residual U-Net backbone while introducing multiplicative modeling on a normalized nonlinear representation, making it more compatible with dense prediction.

\subsection{PU-UNet Architecture}
PU-UNet adopts a residual encoder--decoder architecture with a stem layer, a four-stage encoder, a bottleneck, and a symmetric decoder with skip connections, as illustrated in Fig.~\ref{fig:net}(c). In the default setting, the network has depth $4$ and base width $32$, yielding channel dimensions $32$, $64$, $128$, $256$, and $512$ at relative resolutions $1$, $1/2$, $1/4$, $1/8$, and $1/16$, respectively.

The baseline backbone uses residual-style blocks throughout, and PU-UNet differs only by replacing selected blocks with product-unit residual blocks. In the main configuration, these blocks are inserted into the final encoder block and bottleneck block at $1/16$ resolution, and the first two decoder blocks at $1/8$ and $1/4$ resolution, while all others remain standard residual blocks. This places multiplicative modeling in semantically richer low-resolution stages while preserving additive processing in higher-resolution layers for stable local feature extraction and fine spatial detail. All Group Normalization layers use 8 groups.

\subsection{Training Objective}
For binary segmentation, we optimize a combination of binary cross-entropy (BCE) and focal Tversky loss~\cite{abraham2019novel}. Let $P=\sigma(\hat{Y})$ be the predicted probability map and $Y$ the binary ground-truth mask. The Tversky index~\cite{salehi2017tversky} is defined as
\begin{equation}
\mathrm{TI}(P,Y)=
\frac{\mathrm{TP}+\epsilon}
{\mathrm{TP}+\alpha\,\mathrm{FN}+(1-\alpha)\,\mathrm{FP}+\epsilon},
\label{eq:method_ti}
\end{equation}
where $\mathrm{TP}=\sum (P\odot Y)$, $\mathrm{FP}=\sum (P\odot (1-Y))$, and $\mathrm{FN}=\sum ((1-P)\odot Y)$, and $\odot$ denotes element-wise multiplication. The overall loss is
\begin{equation}
\mathcal{L}
=
\mathcal{L}_{\mathrm{BCE}}(\hat{Y},Y)
+
\left(1-\mathrm{TI}(P,Y)\right)^{\gamma},
\label{eq:method_loss}
\end{equation}
with $\alpha=0.7$ and $\gamma=1.33$, following a common focal Tversky setting~\cite{abraham2019novel}.

\section{Experiments}
\subsection{Datasets and Implementation}
We evaluate PU-UNet on ISIC 2018~\cite{codella2019skin}, Kvasir-SEG~\cite{jha2019kvasir}, and BUSI~\cite{al2020dataset}, all as binary segmentation tasks. ISIC 2018 uses the official training/validation/test split. Kvasir-SEG uses a random $80/10/10$ split, and BUSI uses a stratified $80/10/10$ split with both a main setting including normal images as empty-mask samples and a lesion-only setting excluding normal cases. Random splits were generated with a fixed seed of 42. Images are resized to $352\times352$ for ISIC 2018 and Kvasir-SEG, and to $256\times256$ for BUSI.

Unless otherwise stated, all experiments use the same PU-UNet configuration with base width $32$, depth $4$, SiLU activation, and group normalization. We train with AdamW (initial learning rate $10^{-3}$, weight decay $10^{-4}$), cosine annealing, automatic mixed precision, and batch size $8$ for $500$ epochs on ISIC 2018 and Kvasir-SEG and $300$ epochs on BUSI. Standard spatial augmentations and mild photometric perturbations are used during training.

\subsection{Baselines and Ablation Studies}
We compare PU-UNet with a matched Residual U-Net baseline that uses the same encoder--decoder architecture but replaces all product-unit residual blocks with standard residual blocks.

\emph{Product-unit placement ablation.}
To study where multiplicative modeling is most beneficial, we evaluate two additional placement strategies: (a) \texttt{bottleneck}, where product unit is used only at the bottleneck; and (b) \texttt{all}, where all residual-style blocks are replaced by product-unit residual blocks.

\emph{Stabilization ablation.}
To evaluate the contribution of the proposed stable product unit, we further compare two variants: (a) replacing the smooth positivity mapping (Eq.~\eqref{eq:method_pos}) with a simpler lower-bound truncation (Eq.~\eqref{eq:pu_clip}) used in~\cite{li2025deep}; and (b) removing log-domain clipping, i.e., omitting Eq.~\eqref{eq:method_clip}.

\subsection{Metrics}
We evaluate segmentation performance using Dice and IoU after applying a fixed test-time probability threshold of 0.5. For ISIC 2018, we additionally report thresholded Jaccard at 0.65 to match the official challenge leaderboard, where per-image Jaccard scores below 0.65 are set to zero before averaging. For BUSI with normal images, Dice and IoU are computed on lesion cases only, while normal images are assessed separately by the image-level false-positive rate.

All confidence intervals are 95\% bootstrap confidence intervals estimated from 2,000 resamples. Relative to the Residual U-Net baseline, we also report the absolute performance gain and the corresponding $p$-value from 20,000 paired permutations. Model complexity is measured by parameter count, FLOPs, and inference latency on a single NVIDIA V100 GPU.

\section{Results}
\subsection{Main results}
Tables~\ref{tab:complexity}--\ref{tab:busi_main} show that PU-UNet consistently outperforms the Residual U-Net baseline across all three datasets while maintaining almost identical model complexity. As shown in Table~\ref{tab:complexity}, the two models have nearly the same numbers of parameters, FLOPs, and inference latency at both input resolutions, indicating that the proposed product-unit residual blocks introduce only negligible computational overhead.

Despite this nearly unchanged complexity, PU-UNet achieves clear and statistically significant performance gains on ISIC 2018, Kvasir-SEG, and BUSI (Tables~\ref{tab:isic_kvasir_main} and~\ref{tab:busi_main}). The improvements are consistent across both Dice and IoU. In addition, PU-UNet generally exhibits narrower 95\% confidence intervals than the Residual U-Net baseline, suggesting lower performance variability and more consistent behavior across samples. On BUSI, the proposed model is particularly effective not only in improving lesion segmentation accuracy but also in reducing false-positive predictions on normal cases, which is important for practical deployment.
\begin{table}[h!]
\centering
\caption{Model complexity at the two input resolutions used in our experiments. $\uparrow$ and $\downarrow$ indicate that higher and lower values are better, respectively.}
\label{tab:complexity}
\setlength{\tabcolsep}{5pt}
\renewcommand{\arraystretch}{1.1}
\begin{tabular}{llccc}
\hline
Input size & Method & Parameters (M) $\downarrow$ & FLOPs (G) $\downarrow$ & Latency (ms) $\downarrow$ \\
\hline
\multirow{2}{*}{352$\times$352}
& PU-UNet & 14.393 & 66.727 & 10.213 \\
& Residual U-Net  & 14.391 & 66.725 & 10.073 \\
\hline
\multirow{2}{*}{256$\times$256}
& PU-UNet & 14.393 & 35.294 & 5.610 \\
& Residual U-Net  & 14.391 & 35.292 & 5.530 \\
\hline
\end{tabular}
\end{table}
\begin{table}[h!]
\centering
\setlength{\tabcolsep}{3pt}
\renewcommand{\arraystretch}{1.1}
\caption{Segmentation performance on ISIC 2018 and Kvasir-SEG. 95\% CI denotes the 95\% confidence interval. $p_{Dice/IoU}$ reports the $p$-values for the comparisons between PU-UNet and the Residual U-Net baseline on Dice and IoU, respectively. The best result in each column is highlighted in bold.}
\label{tab:isic_kvasir_main}
\begin{tabular}{lccc}
\hline
Method & Dice (95\% CI) $\uparrow$ & IoU (95\% CI) $\uparrow$ & $p_{Dice/IoU}$ \\
\hline
\multicolumn{4}{l}{\textit{ISIC 2018}}\\
PU-UNet (ours) & \textbf{0.942 [0.937, 0.944]} & \textbf{0.893 [0.887, 0.898]} & -- \\
Residual U-Net (baseline)  & 0.877 [0.868, 0.885] & 0.801 [0.790, 0.812] & -- \\
Improvement      & 0.065 [0.057, 0.071] & 0.092 [0.083, 0.100] & \textbf{$< 0.0001$} \\
\hline
\multicolumn{4}{l}{\textit{Kvasir-SEG}}\\
PU-UNet & \textbf{0.959 [0.956, 0.963]} & \textbf{0.923 [0.916, 0.928]} & -- \\
Residual U-Net  & 0.902 [0.874, 0.925] & 0.839 [0.807, 0.868] & -- \\
Improvement      & 0.057 [0.035, 0.084] & 0.084 [0.056, 0.114] & \textbf{$< 0.0001$} \\
\hline
\end{tabular}
\end{table}
\begin{table}[h!]
\centering
\caption{Lesion segmentation performance and image-level false-positive (FP) rate on BUSI under two evaluation settings: including and excluding normal cases.}
\label{tab:busi_main}
\setlength{\tabcolsep}{2pt}
\renewcommand{\arraystretch}{1.1}
\begin{tabular}{lcccc}
\hline
Method & Lesion Dice (95\% CI) $\uparrow$ & Lesion IoU (95\% CI) $\uparrow$ & FP rate $\downarrow$ & $p_{Dice/IoU}$ \\
\hline
\multicolumn{5}{l}{\textit{Include normal cases}}\\
PU-UNet & \textbf{0.905 [0.876, 0.925]} & \textbf{0.837 [0.804, 0.866]} & \textbf{0.000} & -- \\
Residual U-Net & 0.814 [0.758, 0.867] & 0.733 [0.673, 0.787] & 0.077 & -- \\
Improvement               & 0.091 [0.049, 0.138] & 0.105 [0.067, 0.149] & 0.077 & \textbf{$< 0.0001$} \\
\hline
\multicolumn{5}{l}{\textit{Exclude normal cases}}\\
PU-UNet & \textbf{0.925 [0.904, 0.939]} & \textbf{0.867 [0.841, 0.887]} & -- & -- \\
Residual U-Net & 0.835 [0.787, 0.878] & 0.751 [0.699, 0.800] & -- & -- \\
Improvement               & 0.090 [0.055, 0.131] & 0.116 [0.079, 0.157] & -- & \textbf{$< 0.0001$} \\
\hline
\end{tabular}
\end{table}

\begin{figure}[h!]
    \centering
    \includegraphics[width=0.83\linewidth]{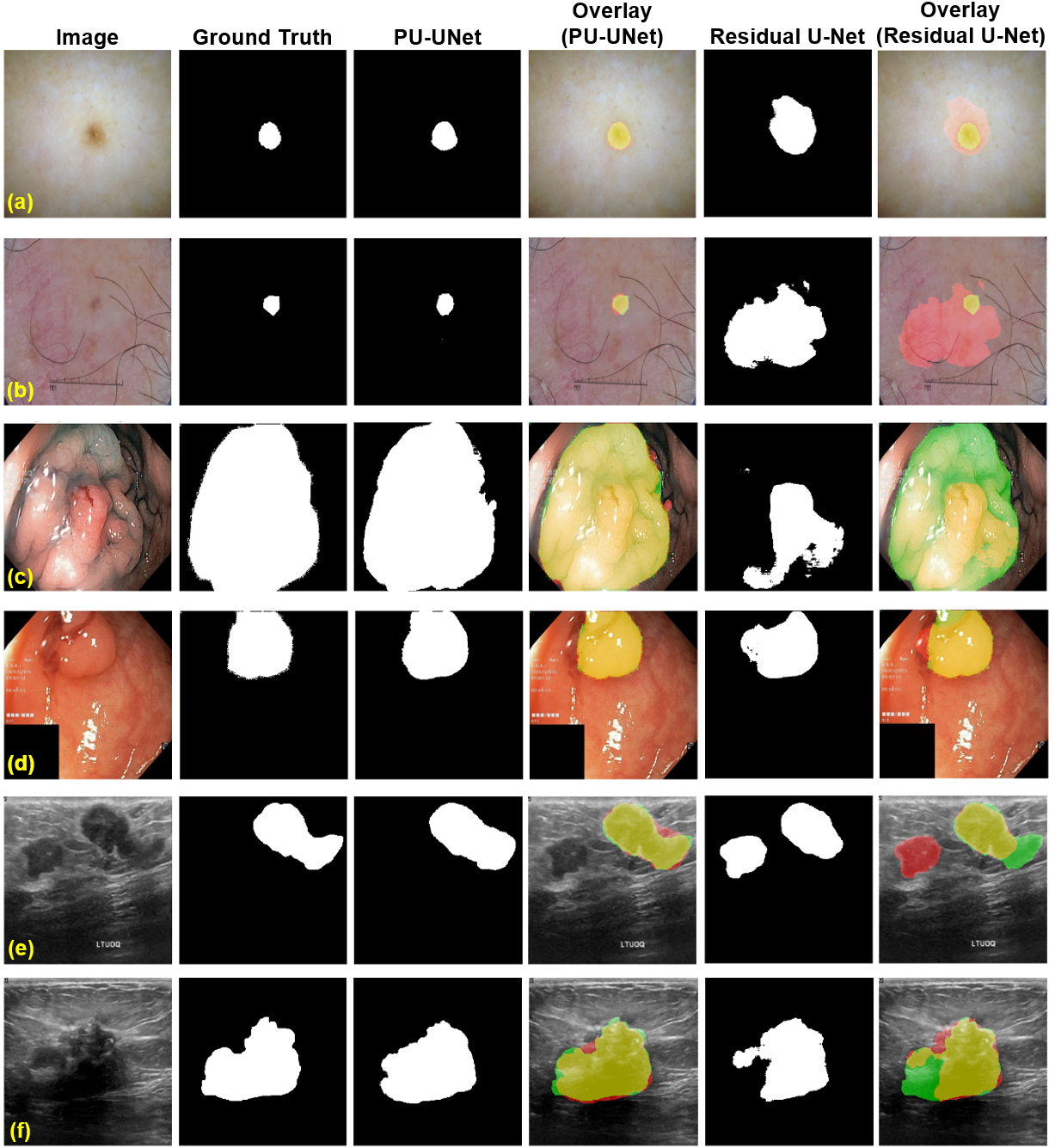}
    \caption{Qualitative segmentation results on three medical image segmentation datasets. Panels (a,b), (c,d), and (e,f) correspond to ISIC 2018, Kvasir-SEG, and BUSI, respectively. In the overlay visualization, green denotes ground truth, red denotes prediction, and yellow denotes their overlap.}
    \label{fig:result}
\end{figure}

Figure~\ref{fig:result} shows representative qualitative comparisons on ISIC 2018, Kvasir-SEG, and BUSI. Across all three datasets, PU-UNet generally produces masks that better match the target extent and preserve object shape, while the Residual U-Net more often exhibits over-segmentation, under-segmentation, or inaccurate localization. The difference is particularly evident on BUSI, where lesion boundaries are less distinct and image quality is more challenging; in these cases, PU-UNet yields more plausible and spatially consistent predictions. These qualitative observations are consistent with the quantitative improvements reported in Tables~\ref{tab:isic_kvasir_main}--\ref{tab:busi_main}.

Table~\ref{tab:related_results_all} demonstrates that PU-UNet is highly competitive with representative published methods across all three datasets. On ISIC 2018, it achieves the best reported IoU/Jaccard and Jaccard@0.65, while matching the best Dice score. On Kvasir-SEG, PU-UNet obtains the best reported IoU together with a highly competitive Dice score. Although UPolySeg reports a higher Dice and ResPVT-B0 also shows strong performance, both use a different 90/10 train/test split rather than our 80/10/10 train/validation/test protocol, so these comparisons are not strictly protocol-matched. On BUSI, PU-UNet achieves the best Dice and IoU in both the main setting including normal cases and the lesion-only setting. Taken together, these results show that selectively introducing stable product-unit transformations into a residual U-Net backbone provides robust and competitive segmentation performance across multiple medical imaging modalities with little additional computational cost.

\begin{table}[h!]
\centering
\caption{Comparison with representative literature on ISIC 2018, Kvasir-SEG, and BUSI. Jaccard@0.65 denotes Jaccard with threshold 0.65.}
\label{tab:related_results_all}
\setlength{\tabcolsep}{4pt}
\begin{tabular}{lccc}
\hline
Method & Dice $\uparrow$ & IoU / Jaccard $^{1}$ $\uparrow$ & Jaccard@0.65 $\uparrow$\\
\hline
\multicolumn{4}{l}{\textit{ISIC 2018}}\\
PU-UNet (ours) & \textbf{0.942} & \textbf{0.893} & \textbf{0.883}\\
MedSAM + AdSTNet~\cite{chen2025blurred} & 0.927 & 0.868 & --\\
MAAU~\cite{le2023anti} & 0.881 & 0.809 & --\\
Hybrid TransUNet~\cite{gulzar2022skin} & 0.898 & -- & -- \\
DilatedSkinNet~\cite{kaur2022automatic} & \textbf{0.942} & 0.891 & --\\
Multipath Fusion Model~\cite{alhudhaif2022novel} & 0.880 & 0.800 & --\\
Mask R-CNN (Official winner)~\cite{codella2019skin} & -- & -- & 0.802 \\
\hline

\multicolumn{4}{l}{\textit{Kvasir-SEG}}\\
PU-UNet (ours) & 0.959 & \textbf{0.923} & --\\
PCA-TransUNet~\cite{chen2026pca} & 0.952  & --  & -- \\
Dual-branch SSM~\cite{chen2025dual} & 0.917  & -- & -- \\
MedSAM + AdSTNet~\cite{chen2025blurred} & 0.873 & 0.781 & --\\
$\mu$-Net~\cite{emon2025integrated} & 0.940 & 0.887 & --\\
ResPVT-B0 $^{2}$~\cite{nachmani2023segmentation} & 0.954 & 0.918 & --\\
UPolySeg $^{2}$~\cite{mohapatra2022upolyseg} & \textbf{0.969} & 0.879 & --\\
A-DenseUNet~\cite{safarov2021denseunet} & 0.909 & 0.862 & --\\
\hline

\multicolumn{4}{l}{\textit{BUSI (include normal cases)}}\\
PU-UNet (ours) & \textbf{0.905} & \textbf{0.837} & --\\
MedSAM + AdSTNet~\cite{chen2025blurred} & 0.849 & 0.751 & --\\
Multi-Task Framework~\cite{aumente2025multi} & 0.769 & 0.625 & --\\
BGRD-TransUNet~\cite{ji2024bgrd} & 0.850 & 0.767 & --\\
\hline

\multicolumn{4}{l}{\textit{BUSI (exclude normal cases)}}\\
PU-UNet (ours) & \textbf{0.925} & \textbf{0.867} & --\\
D-TransUNet~\cite{wan2024d} & 0.893  & -- & --\\
DBU-Net~\cite{pramanik2023dbu} & 0.853 & 0.743 & --\\
\hline
\end{tabular}
\vspace{2mm}
\begin{minipage}{\textwidth}
\footnotesize
\raggedright
$^{1}$ In the standard, unthresholded setting, the Jaccard index is equivalent to IoU.\\
$^{2}$ These results were obtained using a 90/10 train/test split and are therefore not strictly protocol-matched to our setting (80/10/10 train/validation/test).
\end{minipage}
\end{table}

\subsection{Ablation results}
Table~\ref{tab:ablation_pu_placement} summarizes the effects of product-unit residual block placement and stabilization design. All product-unit variants outperform the Residual U-Net baseline, but the low-resolution configuration performs best across all three datasets. The Bottleneck only variant yields smaller gains, and using product-unit residual blocks at all stages is consistently inferior, suggesting that multiplicative modeling is most beneficial in deeper, semantically richer features.

Both stabilization components are also important. Replacing the smooth positivity mapping with a simpler lower-bound truncation reduces performance, while removing log-domain clipping causes a further drop and gradient explosion during training. The degradation is largest on BUSI, where performance even falls below the Bottleneck only variant, indicating that stabilization is particularly important for more challenging ultrasound images.
\begin{table}[t]
\centering
\caption{Ablation study of PU-UNet, including the placement of product-unit residual blocks and the effect of stabilization components. ``w/'' and ``w/o'' denote with and without, respectively.}
\label{tab:ablation_pu_placement}
\setlength{\tabcolsep}{2pt}
\renewcommand{\arraystretch}{1.1}
\begin{tabular}{lccccC{15mm}C{10mm}}
\hline
\multirow{2}{*}{Configuration} 
& \multicolumn{2}{c}{ISIC 2018} 
& \multicolumn{2}{c}{Kvasir-SEG}
& \multicolumn{2}{c}{BUSI (incl. normal)} \\
\cline{2-7}
& Dice $\uparrow$ & IoU $\uparrow$
& Dice $\uparrow$ & IoU $\uparrow$
& Dice $\uparrow$ & IoU $\uparrow$ \\
\hline
\multicolumn{7}{l}{\textit{Effect of product-unit residual block placement}}\\
Residual U-Net (w/o product-unit) & 0.877 & 0.801 & 0.902 & 0.839 & 0.814 & 0.733 \\
Bottleneck only        & 0.908 & 0.862 & 0.933 & 0.852 & 0.839 & 0.761 \\
Low-resolution stages (PU-UNet)  & \textbf{0.942} & \textbf{0.893} & \textbf{0.959} & \textbf{0.923} & \textbf{0.905} & \textbf{0.837} \\
All stages             & 0.930 & 0.872 & 0.939 & 0.864 & 0.857 & 0.792 \\
\hline
\multicolumn{7}{l}{\textit{Effect of stabilization components}}\\
w/ simpler lower-bound truncation & 0.937 & 0.886 & 0.941 & 0.879 & 0.831 & 0.749 \\
w/o log-domain clipping $^{1}$ & 0.933 & 0.878 & 0.935 & 0.863 & 0.826 & 0.735 \\
\hline
\end{tabular}
\vspace{2mm}
\begin{minipage}{\textwidth}
\footnotesize
\raggedright
$^{1}$ This variant exhibited gradient explosion during training.
\end{minipage}
\end{table}

\section{Discussion and Conclusion}
This work shows that stable product units can introduce explicit multiplicative interactions into medical image segmentation with negligible computational overhead. Compared with a matched Residual U-Net baseline, PU-UNet consistently improves Dice and IoU on ISIC 2018, Kvasir-SEG, and BUSI while keeping parameter count, FLOPs, and inference latency nearly unchanged. The reduction of image-level false positives on normal BUSI cases further highlights its practical relevance in challenging ultrasound segmentation.

The ablation results suggest that both placement and stabilization are important. PU blocks are most effective when inserted selectively into low-resolution stages, whereas bottleneck-only insertion provides smaller gains and full-network replacement is less effective. This suggests that multiplicative modeling is better suited to deeper semantic representations than to shallow layers dominated by local texture and edge cues. In addition, replacing the smooth positivity mapping or removing log-domain clipping degrades performance, supporting the importance of stabilization for reliable product-unit learning in the evaluated dense prediction tasks. Together, these findings suggest that the limited use of product units in deeper dense prediction networks may be largely due to practical numerical barriers, and that stable PU design can make multiplicative feature modeling practical in the evaluated U-Net-style medical segmentation setting.

This study is limited to 2D medical image segmentation with a U-Net-style backbone. Future work will evaluate the proposed PU design in more advanced architectures, other imaging modalities such as CT and MRI, 3D volumetric segmentation, and more challenging domain generalization settings.

\subsubsection*{Acknowledgements.}
This research has received funding from the Ministry of Science and Health of Rhineland-Palatinate, Germany, and the Debeka Krankenversicherungsverein a.G. through the Forschungskolleg Data2Health. 
%
%
\bibliographystyle{splncs04}
\bibliography{reference}
\end{document}